# 3D Skull Recognition Using 3D Matching Technique

Hamdan.O.Alanazi, B.B Zaidan, A.A Zaidan

***Abstract*** —Biometrics has become a "hot" area. Governments are funding research programs focused on biometrics. In this paper the problem of person recognition and verification based on a different biometric application has been addressed. The system is based on the 3DSkull recognition using 3D matching technique, in fact this paper present several bio-metric approaches in order of assign the weak point in term of used the biometric from the authorize person and insure the person who access the data is the real person. The feature of the simulate system shows the capability of using 3D matching system as an efficient way to identify the person through his or her skull by match it with database, this technique grantee fast processing with optimizing the false positive and negative as well .

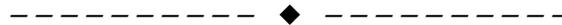

## 1. INTRODUCTION

Biometrics verification is any means by which a person can be uniquely identified by evaluating one or more unique biological traits. Unique identifiers include fingerprints, retina and iris patterns, voice waves, DNA, hand geometry, earlobe geometry and signatures, it has been widely used in the security applications such as Electronic access control, Inmate booking & release/parole ID, Safe & vault security, Elimination of welfare fraud, Information security, ATM [1].

Fingerprint, Iris-pattern and retina-pattern authentication methods are already engaged in some bank automatic teller machines. Voice waveform recognition, a method of verification that has been used for many years with tape recordings in telephone wiretaps, is now being used for access to proprietary databanks in research facilities. Hand geometry is being used in industry to provide physical access to buildings. Facial-recognition technology has been used by law enforcement to pick out individuals in large crowds with considerable reliability [1].

Currently, biometric technology is most identified with border control and transport agencies with both fingerprint and iris scanning being deployed in airports such as Schipol in the Netherlands [2].

―――――――――――――――――――

- *Hamdan.O.Alanazi – Master Studenr, Department of Computer System & Technology, University Malaya, Kuala Lumpur, Malaysia*

- *A. A. Zaidan – PhD Candidate on the Department of Electrical & Computer Engineering , Faculty of Engineering , Multimedia University , Cyberjaya, Malaysia*
- *B. B. Zaidan – PhD Candidate on the Department of Electrical & Computer Engineering / Faculty of Engineering, Multimedia University, Cyberjaya, Malaysia*

However it is also starting to gain prominence in commercial sectors such as financial services, a move being further driven by regulatory compliance across Europe. In the last two years, numerous financial organizations have deployed non-Automated Fingerprint Identification System (AFIS) fingerprint recognition and voice verification to meet FFIEC (Federal Financial Institutions Examination Council) guidelines. Recent analysis from Frost & Sullivan of the world financial biometrics market found that it earned revenues of $117.3 million in 2006, with estimates to reach $2.07 billion in 2013 [1],[2].

Biometric systems are now becoming widely used by many organizations to provide greatest level of security because it more reliable than then password and it represent the user. Many biometric research publications are already done especially related to pattern recognition and digital signal processing issues [2].

Biometrics refers to the measurement of specific attributes or features of the human body, such as the fingerprints, the retinas and irises and even the voice, with the purpose to distinguish that person from others. These characteristics are unique to each individual thus it become an access password to the user. Therefore biometrics is a powerful technology to defend the e-government systems from unauthorized access [3].

Authentication is important concept in security to control over user accessing the system. The most general authentication controls are password and biometric. Password is a common type of first layer defends strategy to determine that a computer user requesting access to a computer system, in fact, the person he claims to be. While password is effective control and the cornerstone of an effective access control system, password is very vulnerable to interception especially when transmitted over a network to authentication machine. Once the password is compromised, the intruders can cause a great harm and significant financial losses [3].

The password system suffers from many drawbacks and unable to positively identify the user. Simplistic passwords are easily anticipated by computer hacker using tools such as password cracker and login spoofing. Once the intruder can obtain the password, the system cannot recognize the true identity of the user and the he has a total access to associated resource. Making the password more complicated sounds good but end users tend to forget difficult string of password. User behavior can actually circumvent password security by failing to choose password wisely, to remember the password, to change the password frequently and to the extreme, to keep the password at secure place [3].

In order to overcome the situation, instead of entering the password alone to gain access to the system, admit needs to prove physical characteristics too. Biometric is capable in differentiating between the authorized user and fraudulent imposter because each and every individuals has unique features that are difficult to be copied. Even though the criminal successfully crack the password, he will not pass the security procedure when the fingerprints check failed [2], [3].

Although there are programs available that can do what is called Steganalysis (Detecting use of Steganography) [4].

The most common use of Steganography is to hide a file inside another file. When information or a file is hidden inside a carrier file, the data is usually encrypted with a password [5]. In this paper the researchers will focus on steganography on digital objects not Steganalysis, more specifically on digital video [6].

## 2. BIOMETRIC APPLICATIONS

### 2.1 Human Iris

Human iris has an extraordinary structure and provides abundant texture information. The spatial patterns that are apparent in the iris are unique to each individual. Individual differences that exist in the development of anatomical structures in the body result in the uniqueness. In particular, the biomedical literature suggests that iris is as distinct as patterns of retinal blood vessels, but an iris image can be more easily obtained than a retina image, Figure 1 shows human iris [7].

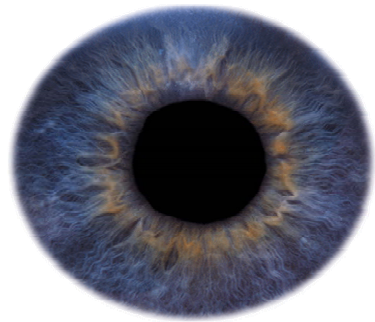

Fig. 1. Human Iris

### 2.2 Human Face Recognition

Human face recognition has been an active research area for the last 20 years. It has many practical applications, such as bankcard identification, access control, mug shots searching, security monitoring, and surveillance systems. Face recognition is used to identify one or more persons from still images or a video image sequence of a scene by comparing input images with faces stored in a database. It is a biometric system that employs automated methods to verify or recognize the identity of a living person based on his physiological characteristic [8].

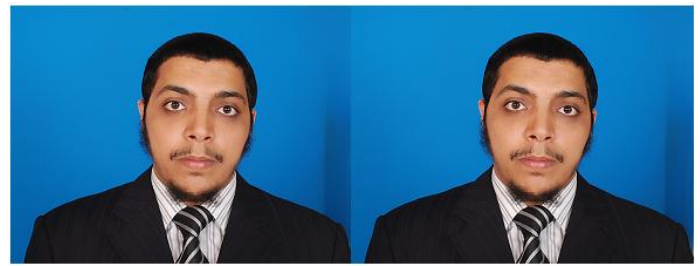

Fig. 2. Face Recognition

### 2.3 Human Voice Biometrics

Human Voice biometrics involves verification of a speaker based on the unique geometry of the speaker's vocal tract such as the vocal tract length, ratio of larynx to sinuses, and the resulting harmonics, pitch, and range. It should be distinguished with speech recognition that does not aim to authenticate a person's identity but understand the meaning of words spoken by a person. Usually, the voice biometric system consists of enrollment process and authentication process. The enrollment process is prior to the authentication process. In enrollment process, the user must be enrolled into the database by creating a reference template of the user's voice-print. In the authentication process, the system authenticates a person's identity by matching a live voice sample with voice-print stored in the system database [9].

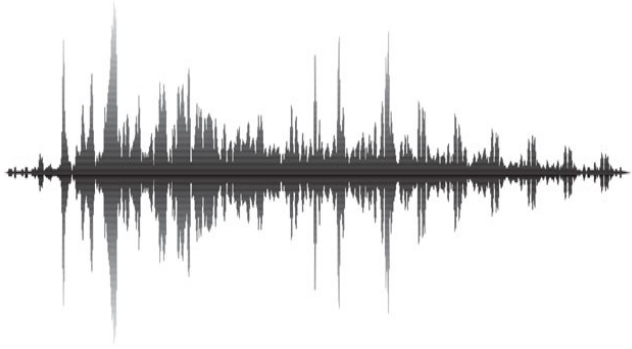

Fig. 3. Voice Recognition

## 2.4 Human Fingerprint

Human fingerprint identification is one of the most important biometric technologies which have drawn a substantial amount of attention recently. A fingerprint is the pattern of ridges and furrows on the surface of a fingertip. Each individual has unique fingerprints. The uniqueness of a fingerprint is exclusively determined by the local ridge characteristics and their relations.

Fingerprint identification is the process of comparing questioned and known friction skin ridge impressions from fingers or palms or even toes to determine if the impressions are from the same finger or palm.

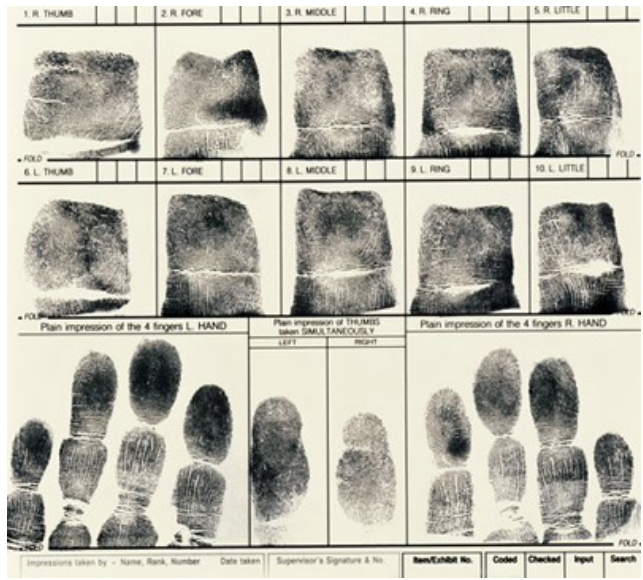

Fig. 4. Fingerprint Recognition

## 2.5 Human Hand Recognition

Human hand recognition is the shape of the hand silhouette as a distinctive personal attribute for an authentication task. Despite the fact that the use of hands as biometric evidence is not very new, and that there are an increasing number of commercial products actually deployed, the documentation in the literature is scarce as compared to other modalities like face or voice. One distinct advantage the hand modality offers is that its imaging conditions are less complex, for example a relatively simple digital camera or flatbed scanner would suffice. Consequently, hand-based biometry is user-friendlier and it is less prone to disturbances and more robust to environmental conditions and to individual anomalies [10].

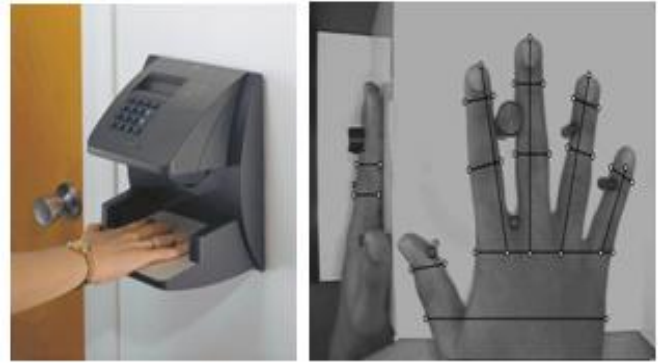

Fig .5. Hand Recognition

## 2.6 Human Blood Vessels

Human blood vessels, in 1935, Simon and Goldstein discovered that the blood vessels (BVs) in the retina are unique to each individual; because of these characteristics, such systems can potentially be used for highly accurate personal recognition. Moreover, since the retina is within the human body, the pattern of retinal BVs remains stable for a long time except in the case of a disease, further, this pattern cannot be replicated by a third person [11].

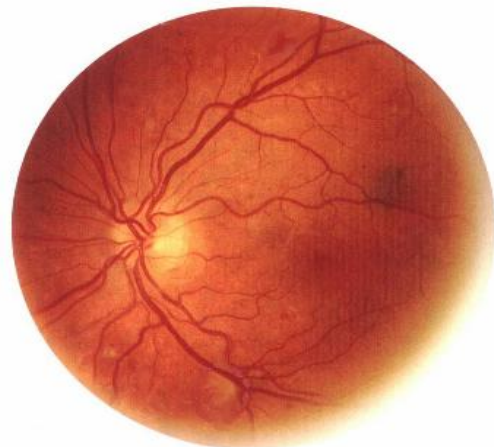

Fig .6. Human Blood Vessels

## 2.7 Ear Biometrics

Ear biometrics can be used as a supplementary source of evidence in identification and recognition systems, for example a system designed for face recognition already includes all the necessary hardware for capturing and computing ear biometrics. Ear biometrics can be used in several scenarios to allow more secure automated access.

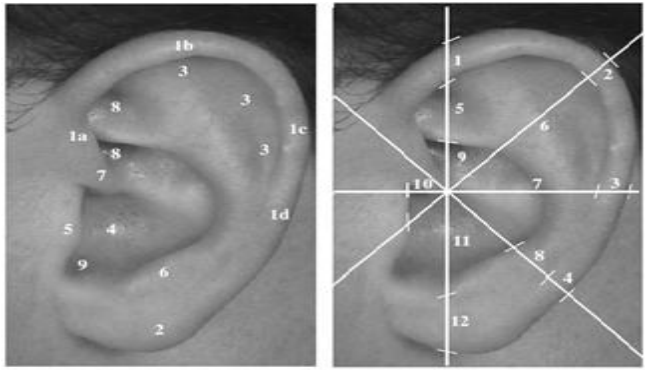

Fig .7. Ear Recognition

## 3. RELATED WORK

Lao et al. [3] perform 3D face recognition using a sparse depth map constructed from stereo images. Iso-luminance contours are used for the stereo matching. Both 2D edges and iso luminance contours are used in finding the irises. In this specific limited sense, this approach is multi-modal. However, there is no separate recognition result from 2D face recognition. Using the iris locations, other feature points are found so that poses standardization can be done. Recognition rates of 87% to 96% are reported using a dataset of ten persons, with four images taken at each of nine poses for each person.

Achermann et al. [4] extend eigenface and hidden Markov model approaches used for 2D face recognition to work with range images. They present results for a dataset of 24 persons, with 10 images per person, and report 100% recognition using an adaptation of the 2D face recognition algorithms.

Tsalakanidou et al. [6] report on multi-modal face recognition using 3D and color images. The use of color rather than simply gray-scale intensity appears to be unique among the multi-modal work surveyed here. Results of experiments using images of 40 persons from the XM2VTS dataset [5] are reported for color images alone, 3D alone, and 3D + color. The recognition algorithm is PCA style matching, plus a combination of the PCA results for the individual color planes and range image. Recognition rates as high as 99% are achieved for the multi-modal algorithm, and multi-modal performance is found to be higher than for either 3D or 2D alone [12],[13].

## 4. SYSTEM FEATURE

This system has choice 3D detection system for the skull and the reason of choosing this Correlation between skull recognition and 3D matching technique is that in this algorithm is to make less process by use the number of sold matching pixel, the figure below shows how the system work with a simple shape such as two square [12],[13].

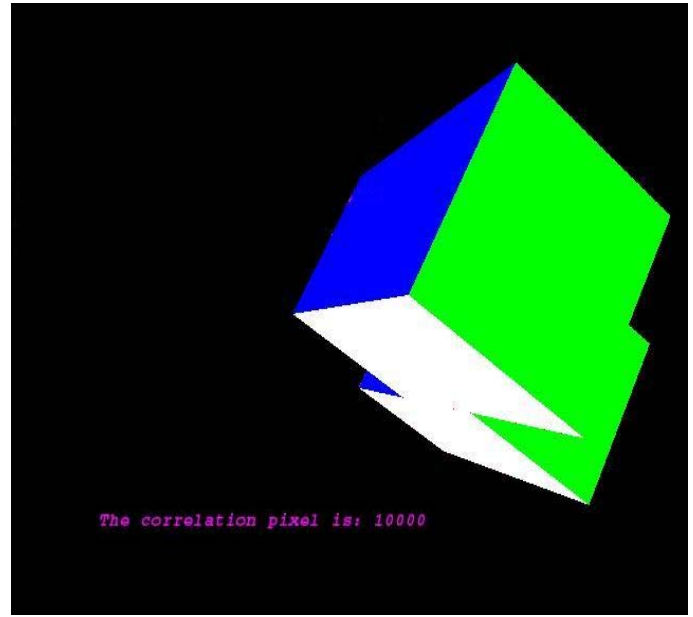

Fig .8. 10000 Pixel Matched Between the Two Square

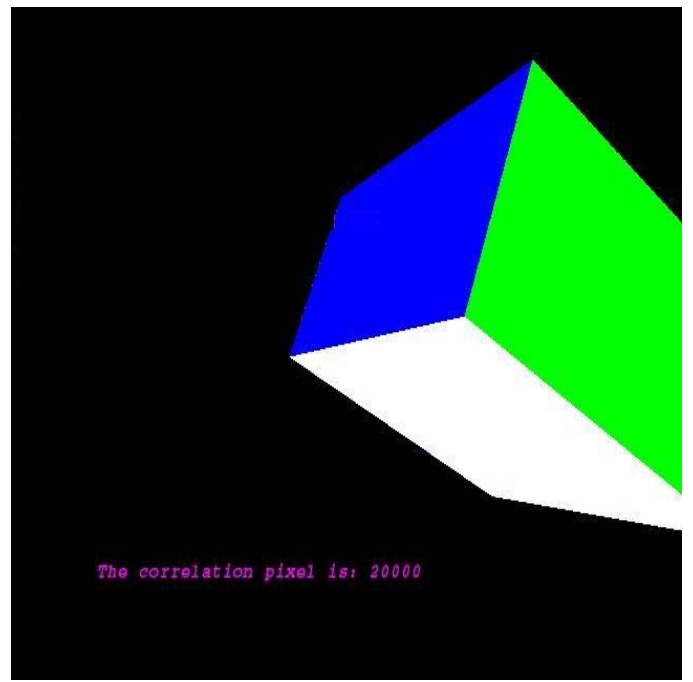

Fig .9. 20000 Pixel Matched Between the Two Square

In the 2nd stage this system required three different points matched afterward calculate the number of matching pixel, if it is same then this skull belong to the same person, if not the skull belong to another person .

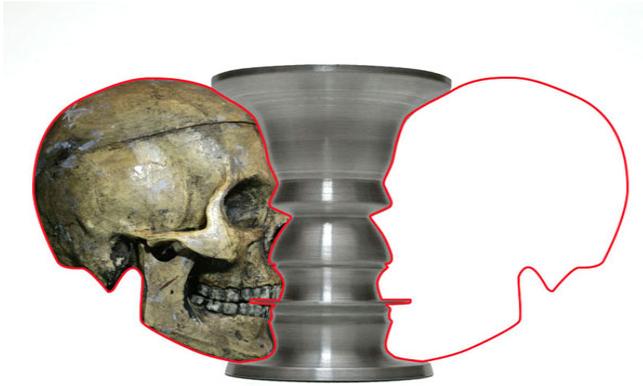

Fig .10. 2D Skull

Fig .9, Fig .10 were a simple test for where in Fig .9 the number of pixels is less than 20000, therefore this squares is not the same, where Fig .10 match all the pixels which means this is the right square [12],[13].

The example of the skull below shows how the points in this system have been chosen.

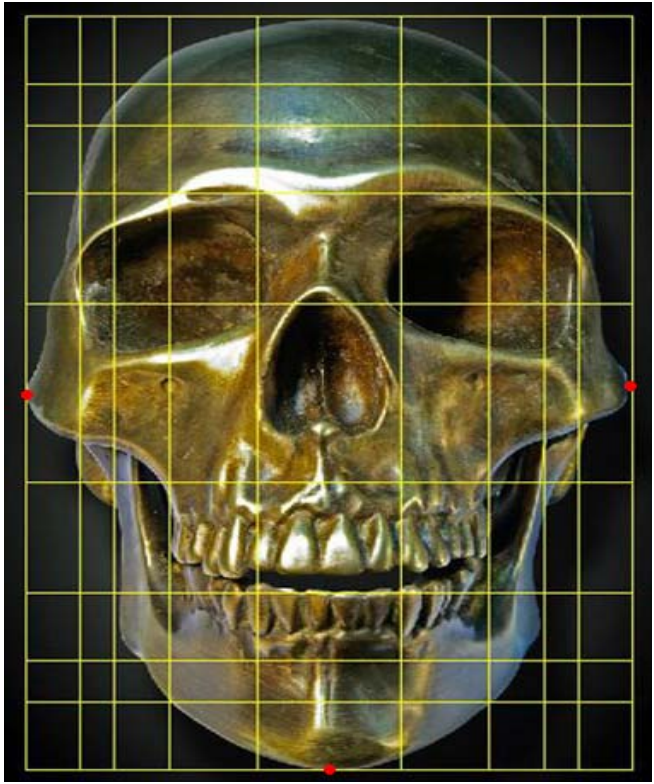

Fig. 11 . The Chosen Point in the Red Color

## 5. CONCLUSION

Reliability in personal authentication is key to the stringent security requirements in many application domains ranging from airport surveillance to electronic banking. Many physiological characteristics of humans. In this paper the author highlight several biometric approaches to determine a wide review about the current bio-metric applications, the simulation of the system and the first result in the simple presentation has been provide, This system has been programmed using java, it has gave a very good result in term of accuracy, time cost and optimizing the false positive and false negative. These entire features have been improved by using a very simple technique. 3D matching with skull has been used in this research project, the reason of using this factors is making auto detection system for the non authorize person using one of the identical stuffs belong to human with a very fast processing as we followed above, the 1st result shows an appreciate result overcome the limitation of the current biometric systems.

## ACKNOWLEDGEMENT

This research was fully supported by "King Saud University", Riyadh, Saudi Arabia. The author would like to acknowledge all workers involved in this project whom had given their support in many ways, aslo he would like to thank in advance Dr. Musaed AL-Jrrah, Dr. Abdullah Alsbail, Dr. Abdullah Alsbait. Dr.Khalid Alhazmi , Dr.Abdullah Al-Afnan, Dr.Ibrahim Al-Dubaian and all the staff in king Saud University especially in Applied Medical Science In "Al-Majmah" for thier unlimited support, without thier notes and suggestion this research would not be appear.

**Hamdan Al-Anazi**: has obtained his bachelor dgree from "King Suad University", Riyadh, Saudi Arabia. He worked as a lecturer at Health College in the Ministry of Health in Saudi Arabia, then he worked as a lecturer at King Saud University in the computer department. Currently he is Master candidate at faculty of Computer Science & Information Technology at University of Malaya in Kuala Lumpur, Malaysia. His research interest on Information Security, cryptography, steganography and digital watermarking, He has contributed to many papers some of them still under reviewer.

**Bilal Bahaa Zaidan:** He obtained his bachelor degree in Mathematics and Computer Application from Saddam University/Baghdad followed by master in data communication and computer network from University of Malaya. He led or member for many funded research projects and He has published more than 50 papers at various international and national conferences and journals, His interest area are Information security (Steganography and Digital watermarking), Network Security (Encryption Methods), Image Processing (Skin Detector), Pattern Recognition, Machine Learning (Neural Network, Fuzzy Logic and Bayesian) Methods and Text Mining and Video Mining. .Currently, he is PhD Candidate on the Department of Electrical & Computer Engineering / Faculty of Engineering / Multimedia University / Cyberjaya, Malaysia. He is members IAENG, CSTA, WASET, and IACSIT. He is reviewer in the (IJSIS, IJCSNS, IJCSN, IJCSE and IJCIIS).

**Aos Alaa Zaidan**: He obtained his 1st Class Bachelor degree in Computer Engineering from university of Technology / Baghdad followed by master in data communication and computer network from University of Malaya. He led or member for many funded research projects and He has published more than 50 papers at various international and national conferences and journals, His interest area are Information security (Steganography and Digital watermarking), Network Security (Encryption Methods), Image Processing (Skin Detector), Pattern Recognition, Machine Learning (Neural Network, Fuzzy Logic and Bayesian) Methods and Text Mining and Video Mining. .Currently, he is PhD Candidate on the Department of Electrical & Computer Engineering / Faculty of Engineering / Multimedia University / Cyberjaya, Malaysia. He is members IAENG, CSTA, WASET, and IACSIT. He is reviewer in the (IJSIS, IJCSNS, IJCSN, IJCSE and IJCIIS).